\pdfoutput=1

\documentclass[11pt]{article}

\usepackage[]{emnlp2021}

\usepackage{times}
\usepackage{latexsym}

\usepackage[T1]{fontenc}

\usepackage[utf8]{inputenc}

\usepackage{microtype}

\usepackage{amsmath}
\usepackage{graphicx}
\usepackage{subcaption}
\usepackage[inline]{enumitem}

\title{Structural analysis of an all-purpose question answering model}

\author{Vincent Micheli \\
  University of Geneva \\
  \texttt{vincent.micheli@unige.ch} \\\And
  Quentin Heinrich \\
  Illuin Technology \\
  \texttt{quentin.heinrich@illuin.tech} \\\AND
  François Fleuret \\
  University of Geneva \\
  \texttt{francois.fleuret@unige.ch} \\\And
  Wacim Belblidia \\
  Illuin Technology \\
  \texttt{wacim.belblidia@illuin.tech}
  }

\begin{document}

\maketitle

\begin{abstract}

Attention is a key component of the now ubiquitous pre-trained language models. By learning to focus on relevant pieces of information, these Transformer-based architectures have proven capable of tackling several tasks at once and sometimes even surpass their single-task counterparts. To better understand this phenomenon, we conduct a structural analysis of a new all-purpose question answering model that we introduce. Surprisingly, this model retains single-task performance even in the absence of a strong transfer effect between tasks. Through attention head importance scoring, we observe that attention heads specialize in a particular task and that some heads are more conducive to learning than others in both the multi-task and single-task settings.

\end{abstract}

\section{Introduction}

Self-supervised learning with deep Transformer-based \citep{attention} networks on a vast text corpus followed by fine-tuning on a specific downstream task \citep{devlin-etal-2019-bert} has become the de facto standard for addressing a myriad of natural language understanding tasks \citep{wang-etal-2018-glue, superglue, rajpurkar-etal-2018-know}. In particular, Question Answering has seen a lot of traction over the past few years with the release of various benchmarks \citep{clark-etal-2019-boolq, rajpurkar-etal-2018-know, kwiatkowski-etal-2019-natural, reddy-etal-2019-coqa}. Building on these models and datasets, we develop a new model capable of answering boolean \citep{clark-etal-2019-boolq} and extractive \citep{rajpurkar-etal-2016-squad, rajpurkar-etal-2018-know} questions such as "Does France have a Prime Minister and a President?" or "When did the 1973 oil crisis begin?". A model of this nature has several advantages:
\begin{enumerate*}[label={(\arabic*)}]
    \item It can be used in numerous use-cases such as information retrieval or conversational agents where boolean or extractive questions may be encountered.
    \item Multi-tasking boolean and extractive question answering alleviates the need for task-specific models and identifying the task at hand.
\end{enumerate*}

We make the surprising observation that even with a limited transfer effect between boolean and extractive question answering, the model is able to reach comparable performance to that of single-task models with the same capacity. To shed light on this finding, we rank the model's attention heads by importance. Indeed, multi-headed attention is essential to producing rich contextualized word representations and has recently been studied in great detail \citep{michel, vig, reif}. By alternately removing attention heads and evaluating the all-purpose model on the downstream question answering tasks, two patterns emerge:
\begin{enumerate*}[label={(\arabic*)}]
\item Head importance is highly task dependent, meaning that some attention heads are critical to carry out a given task and may dampen performance in other cases.
\item A few heads are well-suited for learning in both the single-task and multi-task settings while most heads do not specialize and can be removed at no cost.
\end{enumerate*}

\section{Background: model, datasets}

{\bf RoBERTa} \citep{roberta} is a multi-layer bidirectional Transformer \citep{attention} pre-trained with a standard Masked Language Modeling (MLM) objective. The model improves upon BERT \citep{devlin-etal-2019-bert} by combining several modifications on top of the original optimization procedure and is declined in two architectures: base (12 layers, 768 hidden dimensions, 12 attention heads per layer, 125M parameters) and large (24 layers, 1024 hidden dimensions, 16 attention heads per layer, 355M parameters).

{\bf BoolQ} \citep{clark-etal-2019-boolq} is an open source reading comprehension dataset of 15K naturally occurring boolean questions answered by Wikipedia articles. Given a question and a paragraph found to answer that question, the task consists in answering by yes or no. The authors observed that such questions are particularly challenging and often require complex entailment-like inference.

{\bf SQuAD 2.0} \citep{rajpurkar-etal-2018-know} is a crowd-sourced reading comprehension dataset of extractive questions. Given a paragraph and a question asked about it, the task consists in extracting from the paragraph the span of text answering the question. The dataset contains 151K questions gathered on a set of 442 high-quality Wikipedia articles. It is the extension of SQuAD 1.1 \citep{rajpurkar-etal-2016-squad} with the addition of 53K adversarial questions, i.e. questions that do not have an answer in the associated contexts.

\section{All-purpose question answering}

\subsection{Answering boolean questions}

After a quick step of questions pre-processing (adding question marks and upper casing first letters), byte-level tokenized \citep{sennrich-etal-2016-neural, gpt2} samples are encoded in the following way: questions are concatenated to contexts with a separator token in-between and a sequence representation token is preprended to the whole sequence. After going through the model, the contextualized vector representation of the first token is forwarded to a softmax layer for classification. The training loss is the standard cross-entropy between predicted labels and the ground truth answers.

\subsection{Answering extractive questions}

The encoding procedure is the same as for boolean questions. However, three heads instead of one are put on top of the model:
\begin{enumerate*}[label={(\arabic*)}]
    \item A softmax layer for answer classification, i.e. the paragraph contains an answer to that question or not. Again, the contextualized embedding of the first token is fed to that layer.
    \item Two softmax layers for span classification, i.e. the answer starts or ends with that token or not. These layers slide over the paragraph tokens and for each token predict its likelihood of being the start or the end of the expected answer.
\end{enumerate*}

We denote the parameters of the model as $\theta$, an input sequence as $x$ and the distribution over classes produced by each layer as $f_a(x; \theta)$, $f_s(x; \theta)$ and $f_e(x; \theta)$. The following per-sample loss is minimized during training:
\begin{align*}
L(\theta; x) = \ & l(f_a(x; \theta), y_{a}) \ + \\
& 0.5 \times 1_{\{has\_ans(x)\}} \times l(f_s(x; \theta), y_{s}) \ + \\
& 0.5 \times 1_{\{has\_ans(x)\}} \times l(f_e(x; \theta), y_{e}),
\end{align*}

\noindent 
where $l$ is the cross-entropy, $y_a$, $y_s$ and $y_e$ are the true answer, start token and end token one-hot encoded labels and $1_{\{has\_ans(x)\}}$ is an indicator variable of whether the extractive question is answerable or not. If the question is answerable, then a span of text should be extracted.

\subsection{Answering boolean and extractive questions}

Examples from the two datasets are shuffled together with the usual encoding scheme. The task-specific heads and the training loss are the same as in the extractive setting except that the answer can either be no, yes, adversarial extractive or span extractive. By adversarial extractive we imply an extractive question with no answer while span extractive means an answerable extractive question.

\section{Ranking and masking attention heads}

As a proxy score for attention head importance on a given task, we evaluate the impact of masking that head on development set metrics \citep{michel, mccarley}. 
Masking an attention head means setting its attention matrix to the zero matrix, which in turns sets the tokens representations it produces to zero vectors. 

In the case of BoolQ the evaluation metric is the accuracy while for SQuAD 2.0 it is the F1 score, i.e. the average token overlap between predicted and ground truth answers.

It should be noted that several other head importance metrics could be used in the context of structured masking/pruning \citep{michel, mccarley, movementpruning}. While our proxy score gives a good estimate of a particular head importance given the rest of the model, it does not take into account interactions between heads. Even though combinatorial search would address this issue, it is impractical due to the time consuming evaluation procedure and the total number of heads.

\section{Experiments}

A pre-trained RoBERTa\textsubscript{BASE} is fine-tuned on BoolQ, SQuAD 2.0 and a combination of the two datasets.  The output layers correspond to the setups described in Section 3. For each model, its attention heads are ranked according to the ranking procedure described in Section 4. We also explore a transfer learning approach, where a model fine-tuned on one task is further fine-tuned on the other. Appendix A displays the fine-tuning hyperparameters.

Moreover, in order to assess the difficulty of discriminating between boolean and extractive questions, we fine-tune a tiny (2 layers, 128 hidden dimensions, 2 attention heads) pre-trained BERT model \citep{wellread} on BoolQ and SQuAD's questions. The hyperparameters are the same as for BoolQ.

The experiments described were implemented using Hugging Face’s Transformers library \citep{huggingface} and  were conducted on an NVidia V100 16 GB.

\section{Analysis}

\subsection{All-purpose question answering retains single-task performance}

\begin{table}[h!]
\centering
\begin{tabular}{ll}
\hline \textbf{Task} & \textbf{Evaluation score} \\ \hline
BoolQ & 79.1 Acc. \\
SQuAD & 81 F1 \\
All-purpose & 76 Acc. / 81.4 F1 \\
BoolQ $\rightarrow$ SQuAD & 81.8 F1 \\
SQuAD $\rightarrow$ BoolQ & 81 Acc. \\
\hline
\end{tabular}
\caption{\label{results-table}  RoBERTa\textsubscript{BASE} development set evaluation metrics for each task.}
\end{table}

\begin{table}[h!]
\centering
\begin{tabular}{ll}
\hline \textbf{Task} & \textbf{Evaluation score} \\ \hline
BoolQ & 85.3 Acc. \\
SQuAD & 89.2 F1 \\
All-purpose & 84.5 Acc. / 88.8 F1 \\
\hline
\end{tabular}
\caption{\label{results-table-large} RoBERTa\textsubscript{LARGE} development set evaluation metrics for each task.}
\end{table}

Table \ref{results-table} shows evaluation metrics for the fine-tuned models. We observe that the all-purpose model is close to retaining single-task performance by achieving respectively an accuracy of 76 and an F1 score of 81.4 on the BoolQ and SQuAD development sets. This is a surprising result given that the model has the same architecture as the single-task ones. One hypothesis would be that the two tasks are similar, therefore the model is able to leverage shared knowledge. Indeed, we can observe knowledge transfer across the two tasks to some extent, with an accuracy improvement of 1.9 points on BoolQ following fine-tuning on SQuAD. However, this improvement remains small and comprehensive experiments led by \citet{clark-etal-2019-boolq} did not suggest the existence of a transfer effect between boolean and extractive question answering.

Regarding the all-purpose model scoring slightly lower on BoolQ, we believe there are two main mechanisms at work:
\begin{enumerate*}[label={(\arabic*)}]
    \item The training samples ratio is in favor of SQuAD (13-to-1), resulting in a model biased towards extractive questions. As we will see in the next sections, an important attention head for the BoolQ task actually specialized in answering SQuAD questions.
    \item Fine-tuning being a brittle process \citep{seed}, results should be averaged over multiple runs. As a matter of fact, Table \ref{results-table-large} shows that when repeating the same fine-tuning experiments with RoBERTa\textsubscript{LARGE} instead of RoBERTa\textsubscript{BASE}, no significant performance discrepancy is observed between the all-purpose model and the BoolQ one.
\end{enumerate*}

\subsection{Head importance is highly task dependent}

\begin{figure*}[h!]
  \centering
  \includegraphics[scale=0.27,clip,trim=0 0 90 0,]{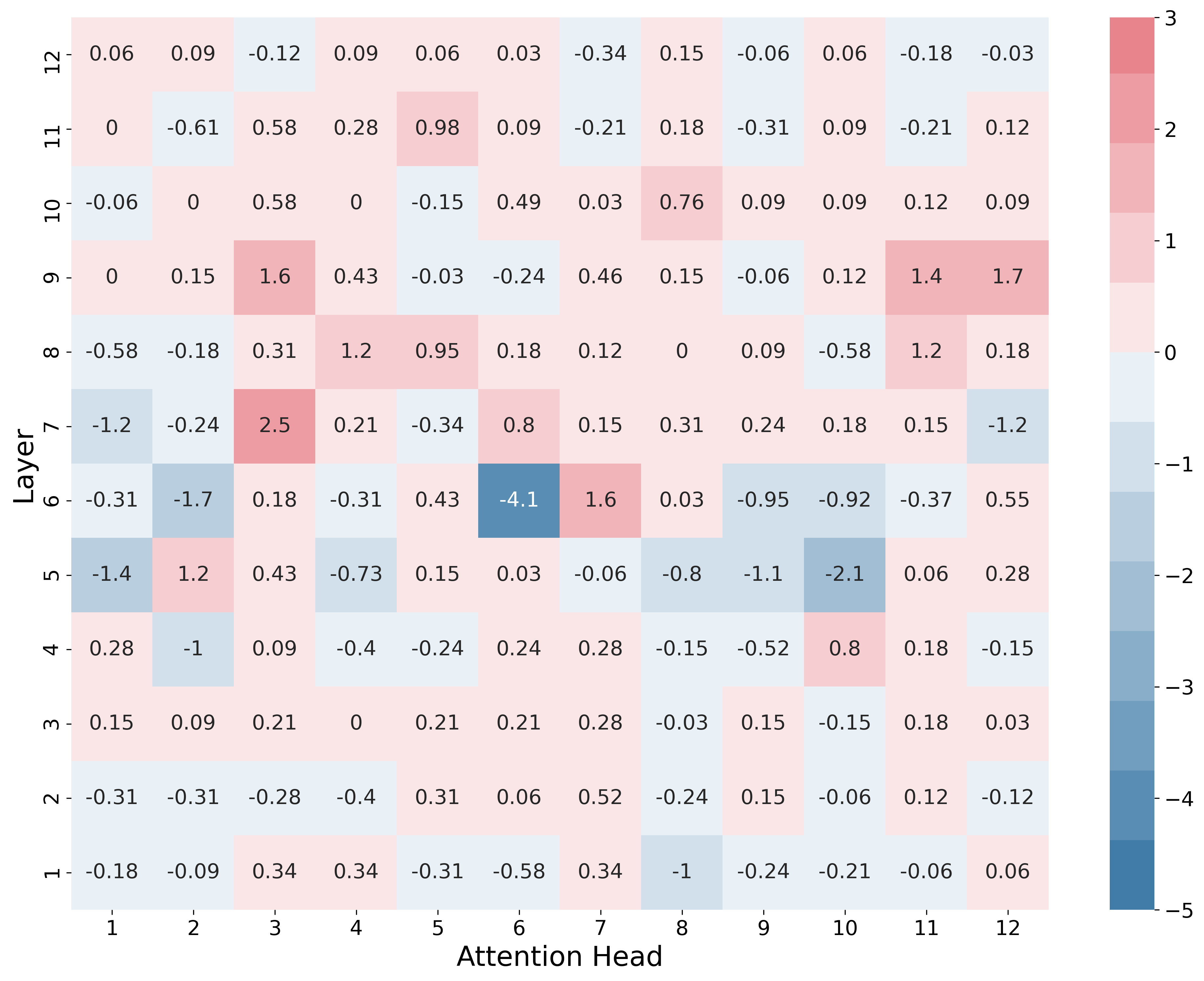}
  \includegraphics[scale=0.27]{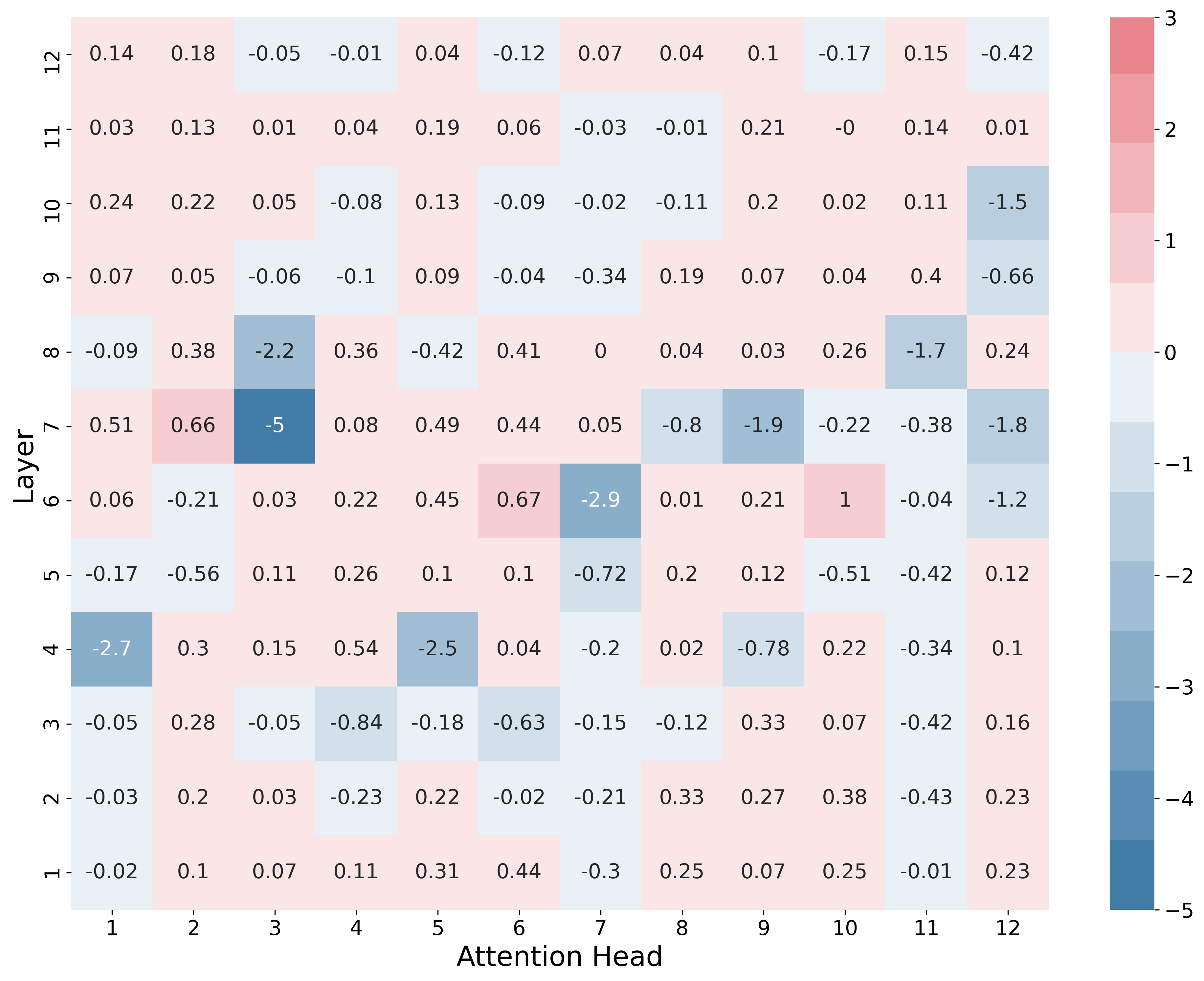}
  \caption{\label{figheadimportanceallpurpose} Change in dev BoolQ Accuracy (left) and dev SQuAD F1 score (right) when masking each head of the all-purpose model.}
\end{figure*}

Figure \ref{figheadimportanceallpurpose} shows head importance scores for the all-purpose model when evaluated on development sets. Interestingly, head importance is highly dependent on the task at hand. In fact, the most important SQuAD head (-5 F1) is the least important BoolQ one (+2.5 accuracy). In fact, its masking allows the model to recover single-task performance when answering boolean questions. Similarly, the most important BoolQ head (-4 accuracy) is the second least important SQuAD one (+0.67 F1). Besides, only one head happens to be in the top 10\% most important heads for both tasks (-1.8 F1, -1.2 accuracy). These results suggest a specialization effect where some heads learn how to answer boolean questions whereas others address extractive questions. In addition, the training of a tiny BERT model reveals that it is perfectly capable to discriminate between boolean and extractive questions. Hence, the all-purpose model should not have to allocate much parameters to classifying questions before answering them.

More generally, most heads can be removed without decreasing evaluation metrics and in many instances removing a head actually results in an increase in performance. These observations echo \citet{michel}'s, except that the change in performance when removing some heads is more pronounced than in their experiments. Moreover, heads located in the intermediate layers have a stronger effect on model performance as illustrated in Table \ref{results-table}, Appendix B and Appendix C.

\subsection{Some heads are more conducive to learning than others}

We also compute head importance scores for single-task models. It is interesting to note that many heads happen to be the most important ones in both the single and multi-task settings. For instance, the most important head stays the same for the all-purpose and SQuAD only models. Furthermore, a detrimental head in the multi-task setting may be important when a model is trained on a single task. Indeed, the all-purpose model’s most detrimental BoolQ head is the third most beneficial head for a BoolQ only model. Again, this shows a specialization effect and it suggests that some heads are more conducive to learning than others. Appendix C further displays an overview of single-task head importance scores.

These results are reminiscent of the lottery ticket hypothesis \citep{lottery}, which states that neural networks contain smaller sub-networks whose initializations are well-suited for learning. Here, the model is pre-trained before fine-tuning. Therefore, some heads may already offer desirable knowledge for those tasks or were better suited for learning at the self-supervised stage. In any case, our coarser-grained analysis reveals that, with a leave-one-out masking approach, a Transformer network is host to two sub-networks of attention heads maintaining single-task performance. 

\section{Related work}

\citet{liu-etal-2019-multi, clark-etal-2019-bam} tackle the problem of multi-task learning with Transformer-based language models. Our multi-task approach deviates from these works as we treat the problem as a single task with a unique set of dedicated classification heads instead of multiple task-specific heads. This has the important benefit of not having to specify the task under consideration.

More recently, UnifiedQA \citep{khashabi-etal-2020-unifiedqa} was proposed to handle various question answering formats. The authors mainly fine-tuned T5 (11B) \citep{t5} models on multiple question answering datasets under the text-to-text paradigm. UnifiedQA achieved performance on-par with single-task models, notably on BoolQ and SQuAD 2.0. In our work, similar results with smaller RoBERTa models motivated a structural analysis to better understand why over-parameterized language models are strong multi-task learners.



\section{Conclusion}

We introduced an all-purpose question answering model, capable of answering both boolean and extractive questions without incurring any significant performance drop nor requiring a larger architecture. Through masking experiments, we showed that a few attentions heads prone to learning specialize in one task in particular. Future works may conduct this structural analysis on new tasks or a greater number of tasks at once. Investigating other head importance metrics taking into account interactions between heads would also further our understanding of learning dynamics.

\section*{Acknowledgements}

We gratefully thank Illuin Technology for its technical support and funding.
We also thank Martin d'Hoffschmidt for his reviewing and helpful discussions.

\bibliography{anthology,custom}

\begin{thebibliography}{25}
\expandafter\ifx\csname natexlab\endcsname\relax\def\natexlab#1{#1}\fi

\bibitem[{Clark et~al.(2019{\natexlab{a}})Clark, Lee, Chang, Kwiatkowski,
  Collins, and Toutanova}]{clark-etal-2019-boolq}
Christopher Clark, Kenton Lee, Ming-Wei Chang, Tom Kwiatkowski, Michael
  Collins, and Kristina Toutanova. 2019{\natexlab{a}}.
\newblock \href {https://doi.org/10.18653/v1/N19-1300} {{B}ool{Q}: Exploring
  the surprising difficulty of natural yes/no questions}.
\newblock In \emph{Proceedings of the 2019 Conference of the North {A}merican
  Chapter of the Association for Computational Linguistics: Human Language
  Technologies, Volume 1 (Long and Short Papers)}, pages 2924--2936,
  Minneapolis, Minnesota. Association for Computational Linguistics.

\bibitem[{Clark et~al.(2019{\natexlab{b}})Clark, Luong, Khandelwal, Manning,
  and Le}]{clark-etal-2019-bam}
Kevin Clark, Minh-Thang Luong, Urvashi Khandelwal, Christopher~D. Manning, and
  Quoc~V. Le. 2019{\natexlab{b}}.
\newblock \href {https://doi.org/10.18653/v1/P19-1595} {{BAM}! born-again
  multi-task networks for natural language understanding}.
\newblock In \emph{Proceedings of the 57th Annual Meeting of the Association
  for Computational Linguistics}, pages 5931--5937, Florence, Italy.
  Association for Computational Linguistics.

\bibitem[{Devlin et~al.(2019)Devlin, Chang, Lee, and
  Toutanova}]{devlin-etal-2019-bert}
Jacob Devlin, Ming-Wei Chang, Kenton Lee, and Kristina Toutanova. 2019.
\newblock \href {https://doi.org/10.18653/v1/N19-1423} {{BERT}: Pre-training of
  deep bidirectional transformers for language understanding}.
\newblock In \emph{Proceedings of the 2019 Conference of the North {A}merican
  Chapter of the Association for Computational Linguistics: Human Language
  Technologies, Volume 1 (Long and Short Papers)}, pages 4171--4186,
  Minneapolis, Minnesota. Association for Computational Linguistics.

\bibitem[{Dodge et~al.(2020)Dodge, Ilharco, Schwartz, Farhadi, Hajishirzi, and
  Smith}]{seed}
Jesse Dodge, Gabriel Ilharco, Roy Schwartz, Ali Farhadi, Hannaneh Hajishirzi,
  and Noah Smith. 2020.
\newblock \href {http://arxiv.org/abs/2002.06305} {Fine-tuning pretrained
  language models: Weight initializations, data orders, and early stopping}.

\bibitem[{Frankle and Carbin(2019)}]{lottery}
Jonathan Frankle and Michael Carbin. 2019.
\newblock \href {https://openreview.net/forum?id=rJl-b3RcF7} {The lottery
  ticket hypothesis: Finding sparse, trainable neural networks}.
\newblock In \emph{International Conference on Learning Representations}.

\bibitem[{Khashabi et~al.(2020)Khashabi, Min, Khot, Sabharwal, Tafjord, Clark,
  and Hajishirzi}]{khashabi-etal-2020-unifiedqa}
Daniel Khashabi, Sewon Min, Tushar Khot, Ashish Sabharwal, Oyvind Tafjord,
  Peter Clark, and Hannaneh Hajishirzi. 2020.
\newblock \href {https://doi.org/10.18653/v1/2020.findings-emnlp.171}
  {{UNIFIEDQA}: Crossing format boundaries with a single {QA} system}.
\newblock In \emph{Findings of the Association for Computational Linguistics:
  EMNLP 2020}, pages 1896--1907, Online. Association for Computational
  Linguistics.

\bibitem[{Kwiatkowski et~al.(2019)Kwiatkowski, Palomaki, Redfield, Collins,
  Parikh, Alberti, Epstein, Polosukhin, Devlin, Lee, Toutanova, Jones, Kelcey,
  Chang, Dai, Uszkoreit, Le, and Petrov}]{kwiatkowski-etal-2019-natural}
Tom Kwiatkowski, Jennimaria Palomaki, Olivia Redfield, Michael Collins, Ankur
  Parikh, Chris Alberti, Danielle Epstein, Illia Polosukhin, Jacob Devlin,
  Kenton Lee, Kristina Toutanova, Llion Jones, Matthew Kelcey, Ming-Wei Chang,
  Andrew~M. Dai, Jakob Uszkoreit, Quoc Le, and Slav Petrov. 2019.
\newblock \href {https://doi.org/10.1162/tacl_a_00276} {Natural questions: A
  benchmark for question answering research}.
\newblock \emph{Transactions of the Association for Computational Linguistics},
  7:452--466.

\bibitem[{Liu et~al.(2019{\natexlab{a}})Liu, He, Chen, and
  Gao}]{liu-etal-2019-multi}
Xiaodong Liu, Pengcheng He, Weizhu Chen, and Jianfeng Gao. 2019{\natexlab{a}}.
\newblock \href {https://doi.org/10.18653/v1/P19-1441} {Multi-task deep neural
  networks for natural language understanding}.
\newblock In \emph{Proceedings of the 57th Annual Meeting of the Association
  for Computational Linguistics}, pages 4487--4496, Florence, Italy.
  Association for Computational Linguistics.

\bibitem[{Liu et~al.(2019{\natexlab{b}})Liu, Ott, Goyal, Du, Joshi, Chen, Levy,
  Lewis, Zettlemoyer, and Stoyanov}]{roberta}
Yinhan Liu, Myle Ott, Naman Goyal, Jingfei Du, Mandar Joshi, Danqi Chen, Omer
  Levy, Mike Lewis, Luke Zettlemoyer, and Veselin Stoyanov. 2019{\natexlab{b}}.
\newblock \href {http://arxiv.org/abs/1907.11692} {Roberta: A robustly
  optimized bert pretraining approach}.

\bibitem[{McCarley et~al.(2020)McCarley, Chakravarti, and Sil}]{mccarley}
J.~S. McCarley, Rishav Chakravarti, and Avirup Sil. 2020.
\newblock \href {http://arxiv.org/abs/1910.06360} {Structured pruning of a
  bert-based question answering model}.

\bibitem[{Michel et~al.(2019)Michel, Levy, and Neubig}]{michel}
Paul Michel, Omer Levy, and Graham Neubig. 2019.
\newblock Are sixteen heads really better than one?
\newblock In \emph{Advances in Neural Information Processing Systems}, pages
  14014--14024.

\bibitem[{Radford et~al.(2019)Radford, Wu, Child, Luan, Amodei, and
  Sutskever}]{gpt2}
Alec Radford, Jeffrey Wu, Rewon Child, David Luan, Dario Amodei, and Ilya
  Sutskever. 2019.
\newblock Language models are unsupervised multitask learners.
\newblock \emph{OpenAI blog}, 1(8):9.

\bibitem[{Raffel et~al.(2020)Raffel, Shazeer, Roberts, Lee, Narang, Matena,
  Zhou, Li, and Liu}]{t5}
Colin Raffel, Noam Shazeer, Adam Roberts, Katherine Lee, Sharan Narang, Michael
  Matena, Yanqi Zhou, Wei Li, and Peter~J. Liu. 2020.
\newblock \href {http://jmlr.org/papers/v21/20-074.html} {Exploring the limits
  of transfer learning with a unified text-to-text transformer}.
\newblock \emph{Journal of Machine Learning Research}, 21(140):1--67.

\bibitem[{Rajpurkar et~al.(2018)Rajpurkar, Jia, and
  Liang}]{rajpurkar-etal-2018-know}
Pranav Rajpurkar, Robin Jia, and Percy Liang. 2018.
\newblock \href {https://doi.org/10.18653/v1/P18-2124} {Know what you don{'}t
  know: Unanswerable questions for {SQ}u{AD}}.
\newblock In \emph{Proceedings of the 56th Annual Meeting of the Association
  for Computational Linguistics (Volume 2: Short Papers)}, pages 784--789,
  Melbourne, Australia. Association for Computational Linguistics.

\bibitem[{Rajpurkar et~al.(2016)Rajpurkar, Zhang, Lopyrev, and
  Liang}]{rajpurkar-etal-2016-squad}
Pranav Rajpurkar, Jian Zhang, Konstantin Lopyrev, and Percy Liang. 2016.
\newblock \href {https://doi.org/10.18653/v1/D16-1264} {{SQ}u{AD}: 100,000+
  questions for machine comprehension of text}.
\newblock In \emph{Proceedings of the 2016 Conference on Empirical Methods in
  Natural Language Processing}, pages 2383--2392, Austin, Texas. Association
  for Computational Linguistics.

\bibitem[{Reddy et~al.(2019)Reddy, Chen, and Manning}]{reddy-etal-2019-coqa}
Siva Reddy, Danqi Chen, and Christopher~D. Manning. 2019.
\newblock \href {https://doi.org/10.1162/tacl_a_00266} {{C}o{QA}: A
  conversational question answering challenge}.
\newblock \emph{Transactions of the Association for Computational Linguistics},
  7:249--266.

\bibitem[{Reif et~al.(2019)Reif, Yuan, Wattenberg, Viegas, Coenen, Pearce, and
  Kim}]{reif}
Emily Reif, Ann Yuan, Martin Wattenberg, Fernanda~B Viegas, Andy Coenen, Adam
  Pearce, and Been Kim. 2019.
\newblock Visualizing and measuring the geometry of bert.
\newblock In \emph{Advances in Neural Information Processing Systems}, pages
  8594--8603.

\bibitem[{Sanh et~al.(2020)Sanh, Wolf, and Rush}]{movementpruning}
Victor Sanh, Thomas Wolf, and Alexander~M. Rush. 2020.
\newblock \href {http://arxiv.org/abs/2005.07683} {Movement pruning: Adaptive
  sparsity by fine-tuning}.

\bibitem[{Sennrich et~al.(2016)Sennrich, Haddow, and
  Birch}]{sennrich-etal-2016-neural}
Rico Sennrich, Barry Haddow, and Alexandra Birch. 2016.
\newblock \href {https://doi.org/10.18653/v1/P16-1162} {Neural machine
  translation of rare words with subword units}.
\newblock In \emph{Proceedings of the 54th Annual Meeting of the Association
  for Computational Linguistics (Volume 1: Long Papers)}, pages 1715--1725,
  Berlin, Germany. Association for Computational Linguistics.

\bibitem[{Turc et~al.(2019)Turc, Chang, Lee, and Toutanova}]{wellread}
Iulia Turc, Ming-Wei Chang, Kenton Lee, and Kristina Toutanova. 2019.
\newblock \href {http://arxiv.org/abs/1908.08962} {Well-read students learn
  better: On the importance of pre-training compact models}.

\bibitem[{Vaswani et~al.(2017)Vaswani, Shazeer, Parmar, Uszkoreit, Jones,
  Gomez, Kaiser, and Polosukhin}]{attention}
Ashish Vaswani, Noam Shazeer, Niki Parmar, Jakob Uszkoreit, Llion Jones,
  Aidan~N Gomez, {\L}ukasz Kaiser, and Illia Polosukhin. 2017.
\newblock Attention is all you need.
\newblock In \emph{Advances in neural information processing systems}, pages
  5998--6008.

\bibitem[{Vig(2019)}]{vig}
Jesse Vig. 2019.
\newblock \href {http://arxiv.org/abs/1906.05714} {A multiscale visualization
  of attention in the transformer model}.

\bibitem[{Wang et~al.(2019)Wang, Pruksachatkun, Nangia, Singh, Michael, Hill,
  Levy, and Bowman}]{superglue}
Alex Wang, Yada Pruksachatkun, Nikita Nangia, Amanpreet Singh, Julian Michael,
  Felix Hill, Omer Levy, and Samuel Bowman. 2019.
\newblock Superglue: A stickier benchmark for general-purpose language
  understanding systems.
\newblock In \emph{Advances in Neural Information Processing Systems}, pages
  3266--3280.

\bibitem[{Wang et~al.(2018)Wang, Singh, Michael, Hill, Levy, and
  Bowman}]{wang-etal-2018-glue}
Alex Wang, Amanpreet Singh, Julian Michael, Felix Hill, Omer Levy, and Samuel
  Bowman. 2018.
\newblock \href {https://doi.org/10.18653/v1/W18-5446} {{GLUE}: A multi-task
  benchmark and analysis platform for natural language understanding}.
\newblock In \emph{Proceedings of the 2018 {EMNLP} Workshop {B}lackbox{NLP}:
  Analyzing and Interpreting Neural Networks for {NLP}}, pages 353--355,
  Brussels, Belgium. Association for Computational Linguistics.

\bibitem[{Wolf et~al.(2020)Wolf, Debut, Sanh, Chaumond, Delangue, Moi, Cistac,
  Rault, Louf, Funtowicz, Davison, Shleifer, von Platen, Ma, Jernite, Plu, Xu,
  Scao, Gugger, Drame, Lhoest, and Rush}]{huggingface}
Thomas Wolf, Lysandre Debut, Victor Sanh, Julien Chaumond, Clement Delangue,
  Anthony Moi, Pierric Cistac, Tim Rault, Rémi Louf, Morgan Funtowicz, Joe
  Davison, Sam Shleifer, Patrick von Platen, Clara Ma, Yacine Jernite, Julien
  Plu, Canwen Xu, Teven~Le Scao, Sylvain Gugger, Mariama Drame, Quentin Lhoest,
  and Alexander~M. Rush. 2020.
\newblock \href {http://arxiv.org/abs/1910.03771} {Huggingface's transformers:
  State-of-the-art natural language processing}.

\end{thebibliography}
\bibliographystyle{acl_natbib}

\onecolumn
\appendix

\section{Hyperparameters}

\begin{table*}[h!]
\centering
\tabcolsep=0.11cm
\begin{tabular}{llll}
\hline \textbf{Parameter} & \textbf{BoolQ} & \textbf{SQuAD} & \textbf{All-purpose} \\ \hline
Epochs & 5 & 3 & 5 \\
Warmup ratio & 0.0 & 0.06 & 0.06 \\
Batch size & 32 & 16 & 16 \\
Learning rate & 1e-5 & 1.5e-5 & 1.5e-5 \\
Adam $\beta_1$ & 0.9 & 0.9 & 0.9 \\
Adam $\beta_2$ & 0.999 & 0.999 & 0.999 \\
Gradient norm & 1.0 & 1.0 & 1.0 \\
Dropout & 0.1 & 0.1 & 0.1 \\
Sequence length & 256 & 384 & 384 \\
\hline
\end{tabular}
\caption{\label{hyperparameters-table} Fine-tuning hyperparameters.}
\end{table*}

\clearpage

\section{Layer-wise head importance scores}
\label{sec:appendixC}

\begin{figure*}[h]
  \centering
  \includegraphics[scale=0.25]{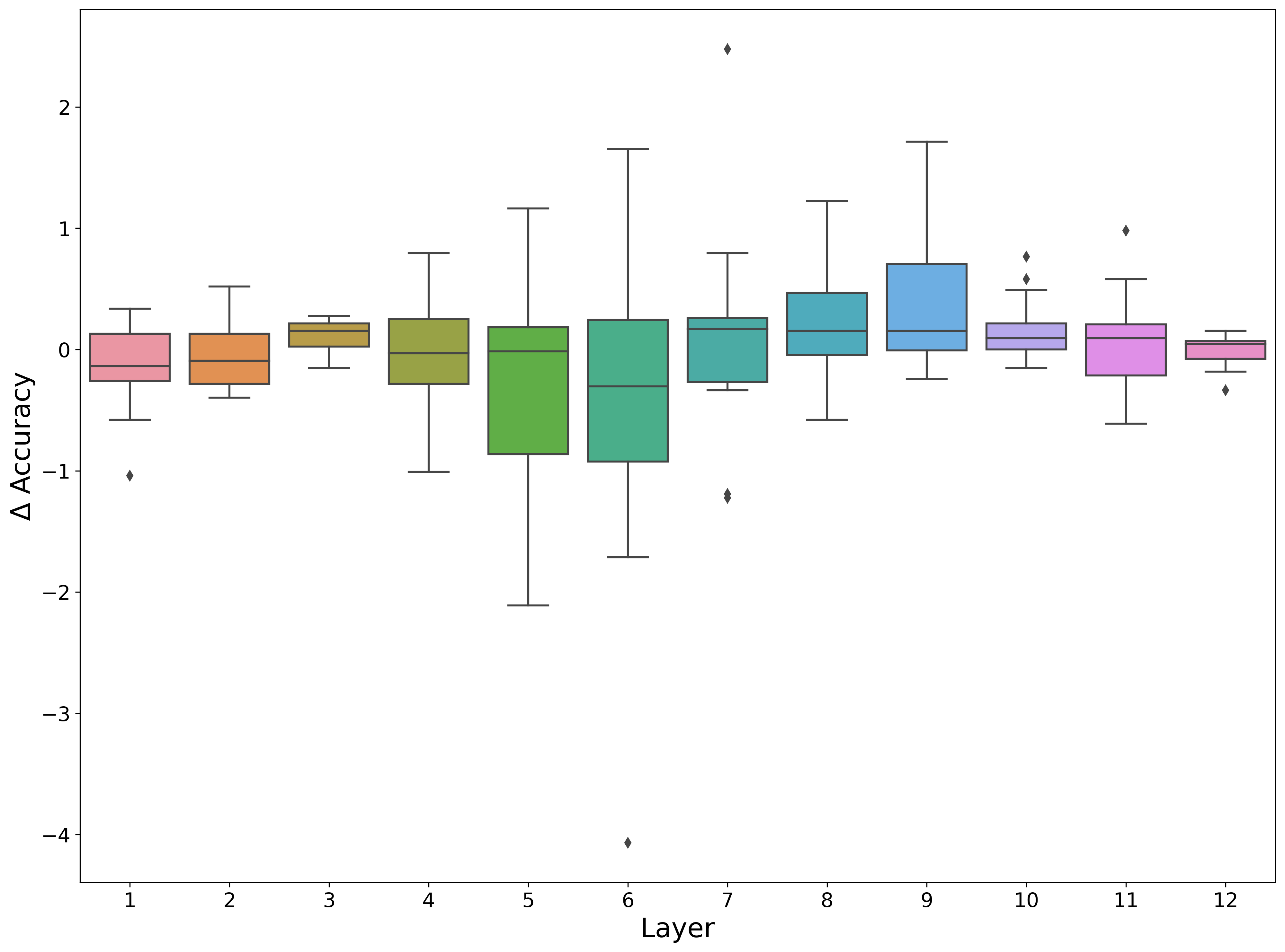}
  \caption{\label{figboxplotboolqmulti} Layer-wise change in dev BoolQ accuracy when masking each head of the all-purpose model.}
\end{figure*}

\begin{figure*}[h]
  \centering
  \includegraphics[scale=0.25]{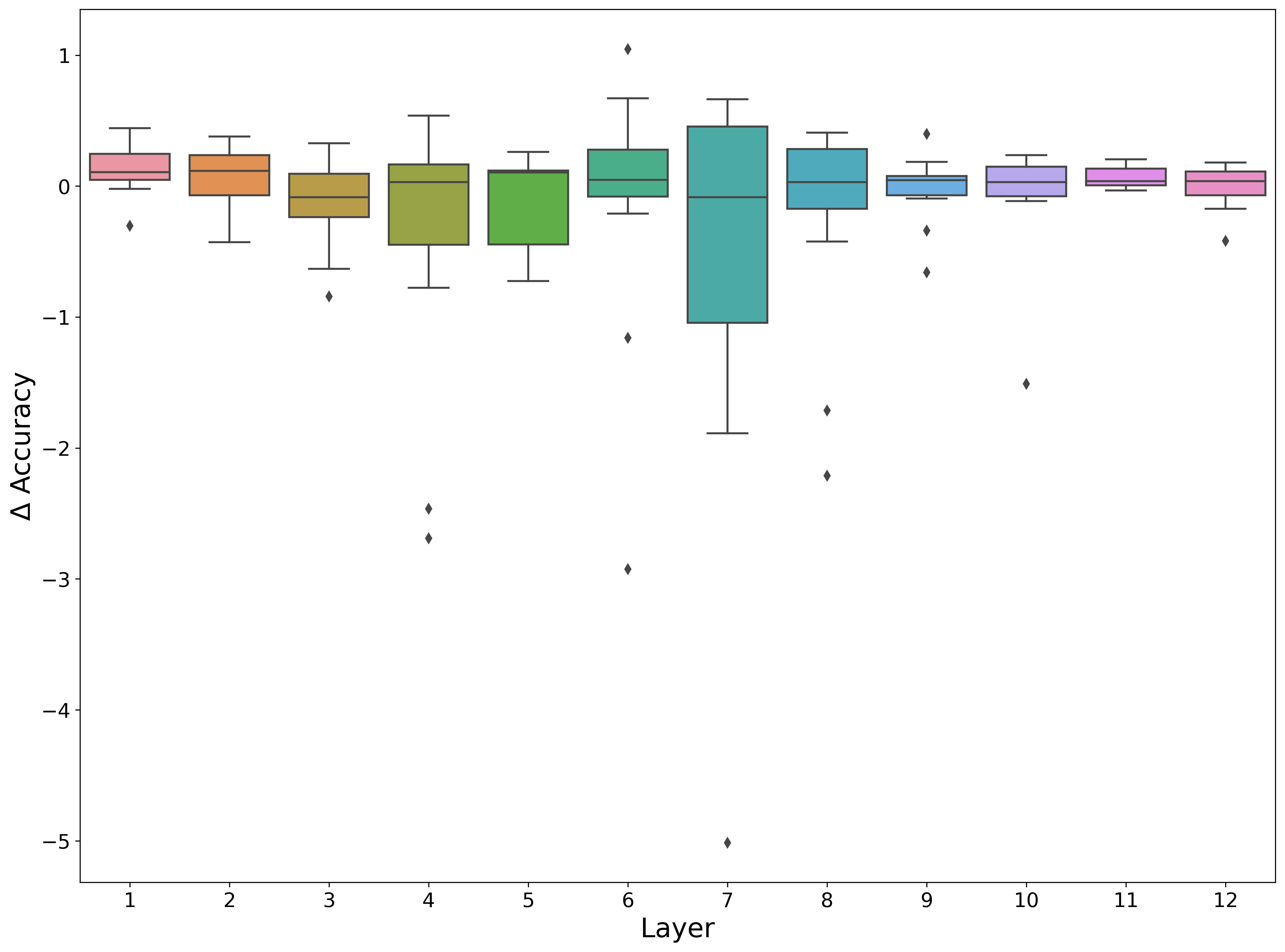}
  \caption{\label{figboxplotsquadmulti} Layer-wise change in dev SQuAD F1 when masking each head of the all-purpose model.}
\end{figure*}

\begin{figure*}[h]
  \centering
  \includegraphics[scale=0.25]{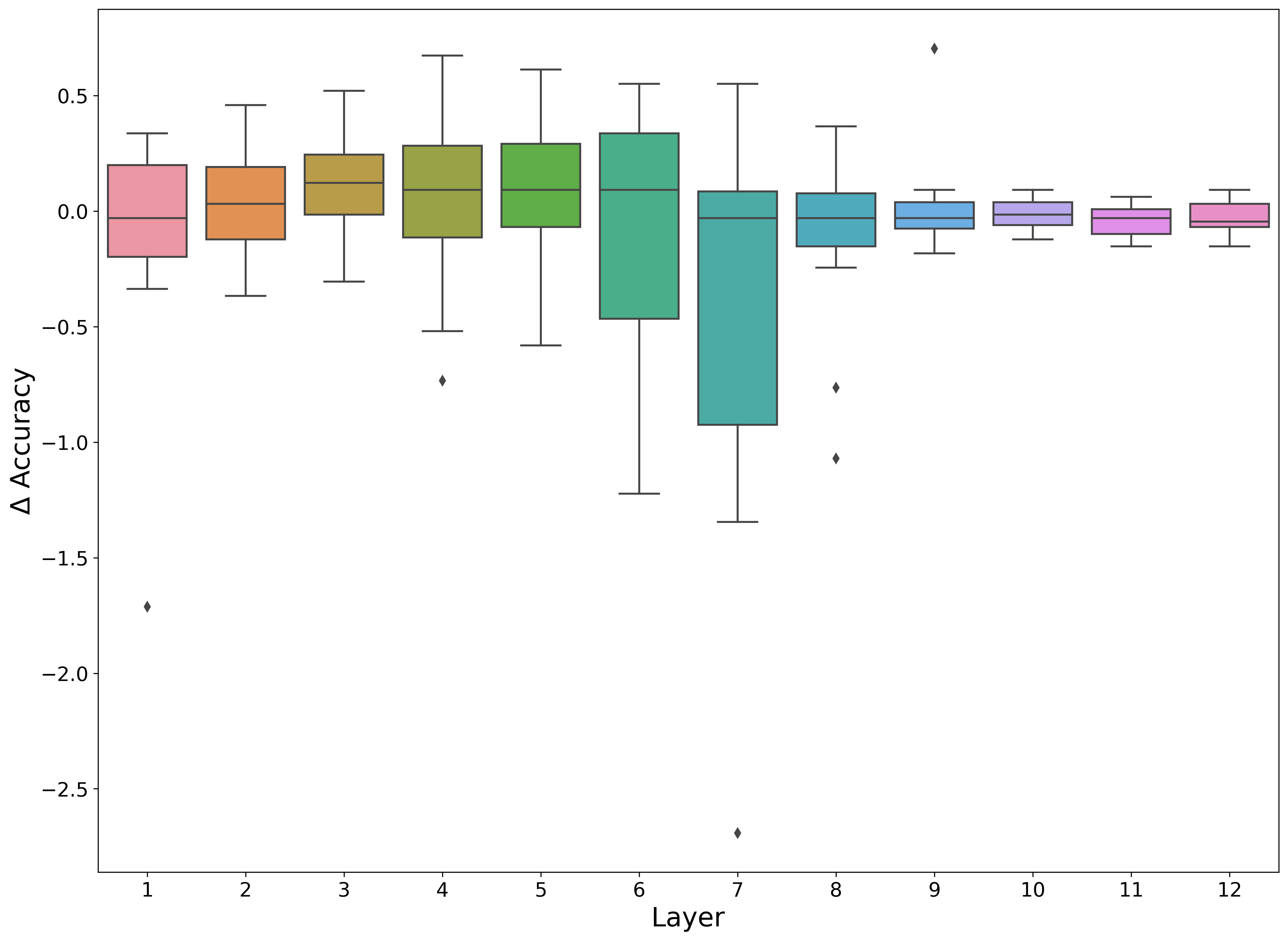}
  \caption{\label{figboxplotboolqsingle} Layer-wise change in dev BoolQ accuracy when masking each head of the BoolQ only model.}
\end{figure*}

\begin{figure*}[h]
  \centering
  \includegraphics[scale=0.25]{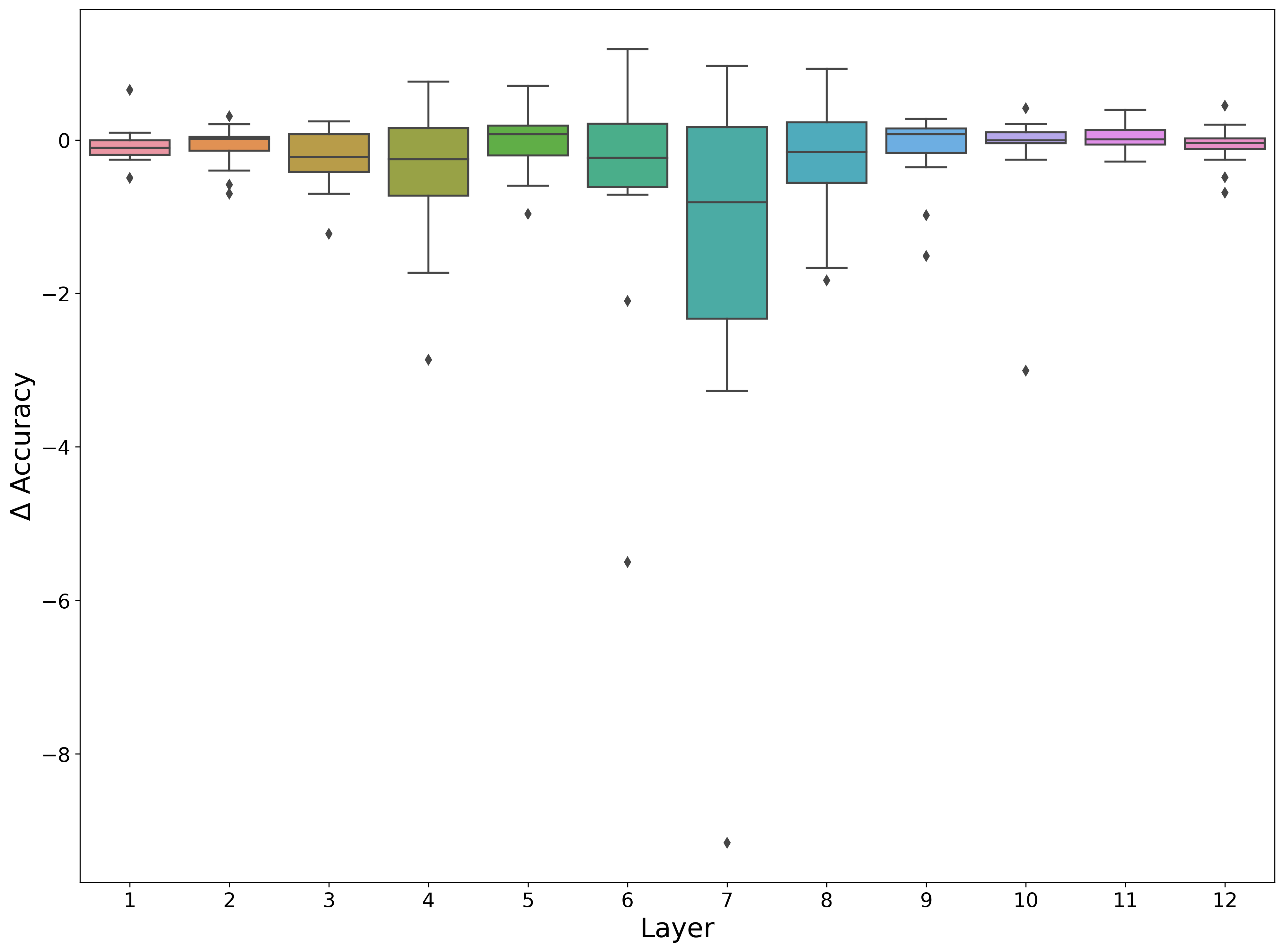}
  \caption{\label{figboxplotsquadsingle} Layer-wise change in dev SQuAD F1 when masking each head of the SQuAD only model.}
\end{figure*}

\clearpage

\section{Single-task head importance scores}

\begin{figure*}[h]
  \centering
  \includegraphics[scale=0.27,clip,trim=0 0 90 0,]{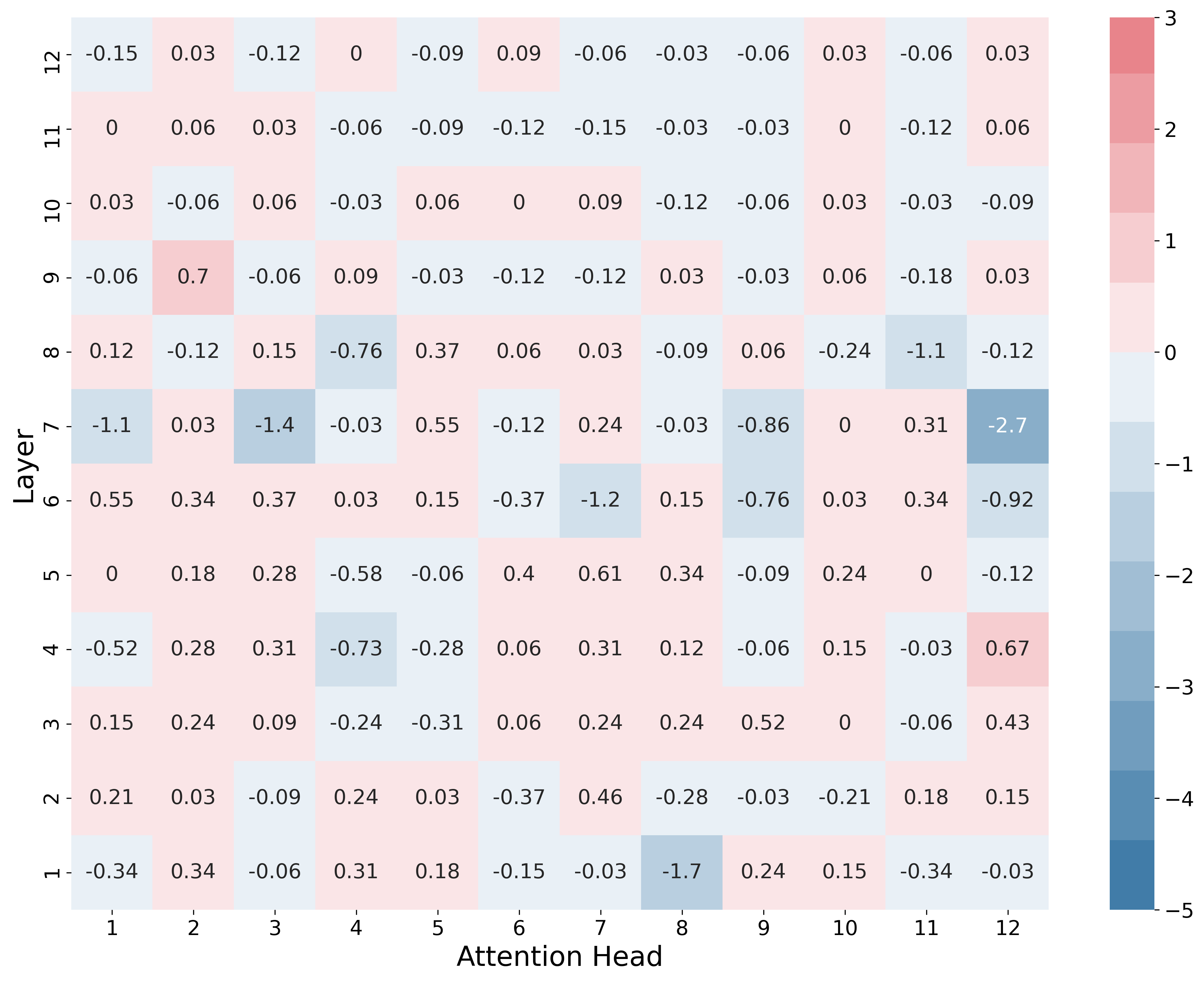}
  \includegraphics[scale=0.27]{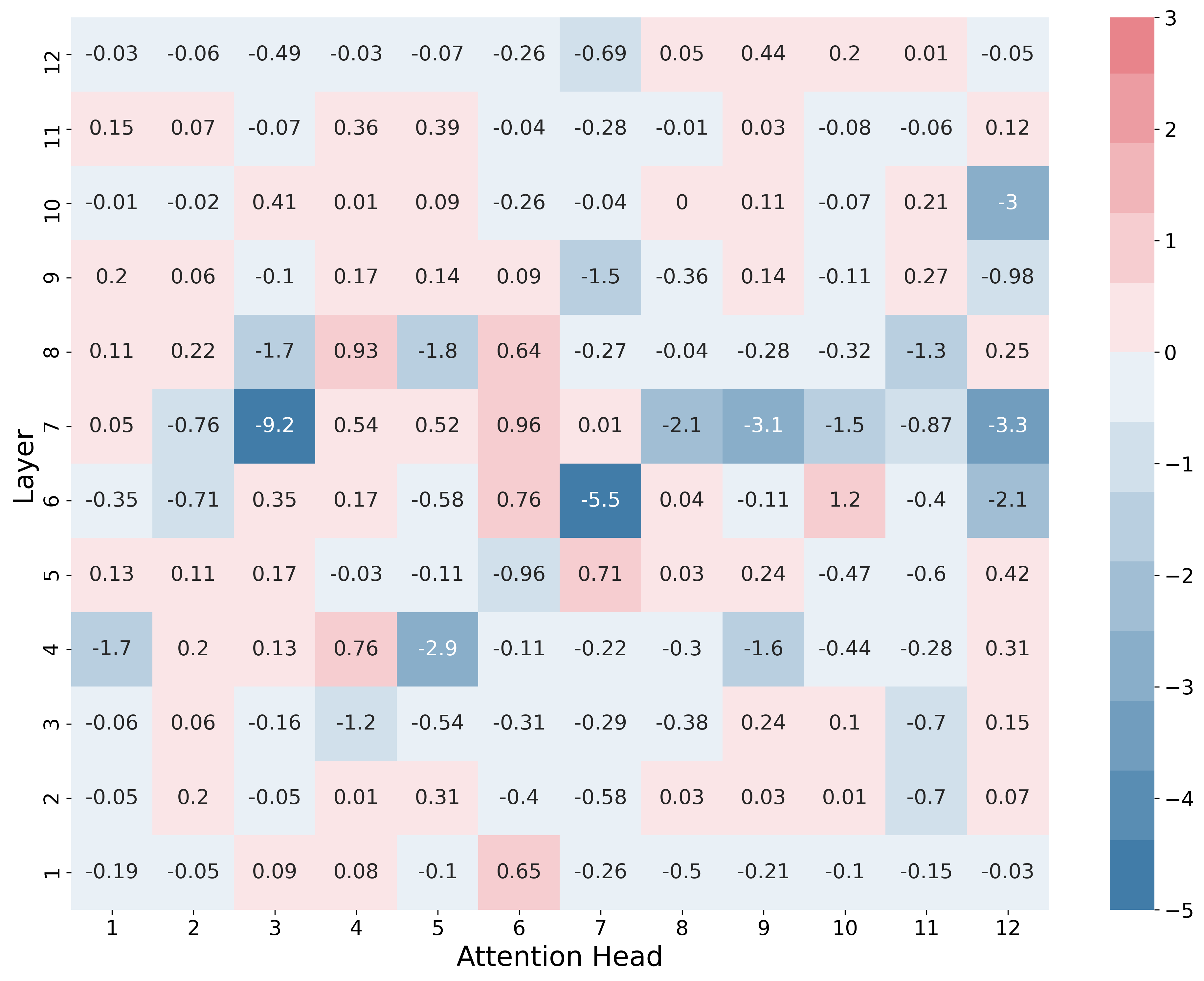}
  \caption{\label{boxplotheadimportanceallpurpose} Change in dev BoolQ Accuracy (left) and dev SQuAD F1 score (right) when masking each head of the all-purpose model.}
\end{figure*}

\clearpage

\section{All-purpose question answering example}

\noindent\makebox[\textwidth][c]{%
\resizebox{0.85\textwidth}{!}{%
\fbox{\begin{minipage}{\textwidth}
    
    \textbf{Paragraph}: Super Bowl 50 was an American football game to determine the champion of the National Football League(NFL) for the 2015 season. The American Football Conference (AFC) champion Denver Broncos defeated the National Football Conference (NFC) champion Carolina Panthers 24–10 to earn their third Super Bowl title. The game was played on February 7, 2016, at Levi's Stadium in the San Francisco Bay Area at Santa Clara, California. As this was the 50th Super Bowl, the league emphasized the 'golden anniversary' with various
gold-themed initiatives, as well as temporarily suspending the tradition of naming each Super Bowl game with Roman numerals (under which the game would have been known as 'Super Bowl L'), so that the logo could prominently feature the Arabic numerals 50.
    
    \vspace{0.1in}
    
    \textbf{Question: }Did the Denver Broncos win the Super Bowl 50?\\
    \textbf{Answer}: Yes

    \vspace{0.1in}
    
    \textbf{Question: }Did the Carolina Panthers win the Super Bowl 50?\\
    \textbf{Answer}: No
    
    \vspace{0.1in}
    
    \textbf{Question: }When did Super Bowl 50 take place?\\
    \textbf{Answer}: Extractive, \textbf{Extracted span}: February 7, 2016
    
    \vspace{0.1in}
    
    \textbf{Question: }Which NFL team won Super Bowl 50?\\
    \textbf{Answer}: Extractive, \textbf{Extracted span}: Denver Broncos
    
    \vspace{0.1in}
    
    \textbf{Question: }Who was the Bronco's coach?\\
    \textbf{Answer}: Adversarial extractive (unknown)

    \vspace{0.1in}
    
    \textbf{Question: }Who did the Broncos defeat in the qualifications?\\
    \textbf{Answer}: Adversarial extractive (unknown)
  
\end{minipage}}}}

\end{document}